%% file: main.tex
\documentclass[sigconf, screen=false, review=false, anonymous=false, balance=false]{acmart}
\usepackage{graphicx}
\usepackage{float}
\usepackage{caption}
\usepackage{subcaption}
\usepackage{svg}
\usepackage{enumitem}

\input{math_commands.tex}

\usepackage{hyperref}
\usepackage{url}
\usepackage{chngpage}

\newif\ifNota
\Notatrue

\AtBeginDocument{%
  \providecommand\BibTeX{{%
    \normalfont B\kern-0.5em{\scshape i\kern-0.25em b}\kern-0.8em\TeX}}}

\begin{document}

\setcopyright{acmcopyright}
\copyrightyear{2023}
\acmYear{2023}
\acmDOI{XXXXXXX.XXXXXXX}

\acmConference[arXiv]{pre-print arXiv}{July 2023}{}
%
%
\acmBooktitle{pre-print ArXiv, July, 2023} 

\title{The GANfather: Controllable generation of malicious activity to improve defence systems}

\author{Ricardo Ribeiro Pereira}
\authornote{Corresponding author, ricardo.ribeiro@feedzai.com}

\iftrue
\affiliation{%
  \institution{Feedzai / Porto University}
  \country{Portugal}
}
\author{Jacopo Bono}
\affiliation{%
  \institution{Feedzai}
  \country{Portugal}
}
\author{João Tiago Ascensão}
\affiliation{%
  \institution{Feedzai}
  \country{Portugal}
}
\author{David Aparício}
\affiliation{%
  \institution{Porto University}
  \country{Portugal}
}
\author{Pedro Ribeiro}
\affiliation{%
  \institution{Porto University}
  \country{Portugal}
}
\author{Pedro Bizarro}
\affiliation{%
  \institution{Feedzai}
  \country{Portugal}
}
\authornote{Corresponding author, pedro.bizarro@feedzai.com}
\fi

\begin{abstract}
Machine learning methods to aid defence systems in detecting malicious activity typically rely on labelled data.
In some domains, such labelled data is unavailable or incomplete.
In practice this can lead to low detection rates and high false positive rates, which characterise for example anti-money laundering systems.
In fact, it is estimated that 1.7--4 trillion euros are laundered annually and go undetected.
We propose \textit{The GANfather}, a method to generate samples with properties of malicious activity, without label requirements.
We propose to reward the generation of malicious samples by introducing an extra objective to the typical Generative Adversarial Networks (GANs) loss.
Ultimately, our goal is to enhance the detection of illicit activity using the discriminator network as a novel and robust defence system.
Optionally, we may encourage the generator to bypass pre-existing detection systems.
This setup then reveals defensive weaknesses for the discriminator to correct.
We evaluate our method in two real-world use cases, money laundering and recommendation systems.
In the former, our method moves cumulative amounts close to 350 thousand dollars through a network of accounts without being detected by an existing system.
In the latter, we recommend the target item to a broad user base with as few as 30 synthetic attackers.
In both cases, we train a new defence system to capture the synthetic attacks.
\end{abstract}

\maketitle

\input{01_intro}

\input{02_method}

\input{03_results}

\input{04_related}

\input{05_conclusion}

\bibliography{ref}
\bibliographystyle{ref}


\end{document}

%% file: math_commands.tex
\usepackage{amsmath,amsfonts,bm}

\def\eqref#1{equation~\ref{#1}}

\def\1{\bm{1}}

\DeclareMathAlphabet{\mathsfit}{\encodingdefault}{\sfdefault}{m}{sl}
\SetMathAlphabet{\mathsfit}{bold}{\encodingdefault}{\sfdefault}{bx}{n}



%% file: 01_intro.tex
\begin{figure}[ht]
    \centering
    \includegraphics[width=\linewidth]{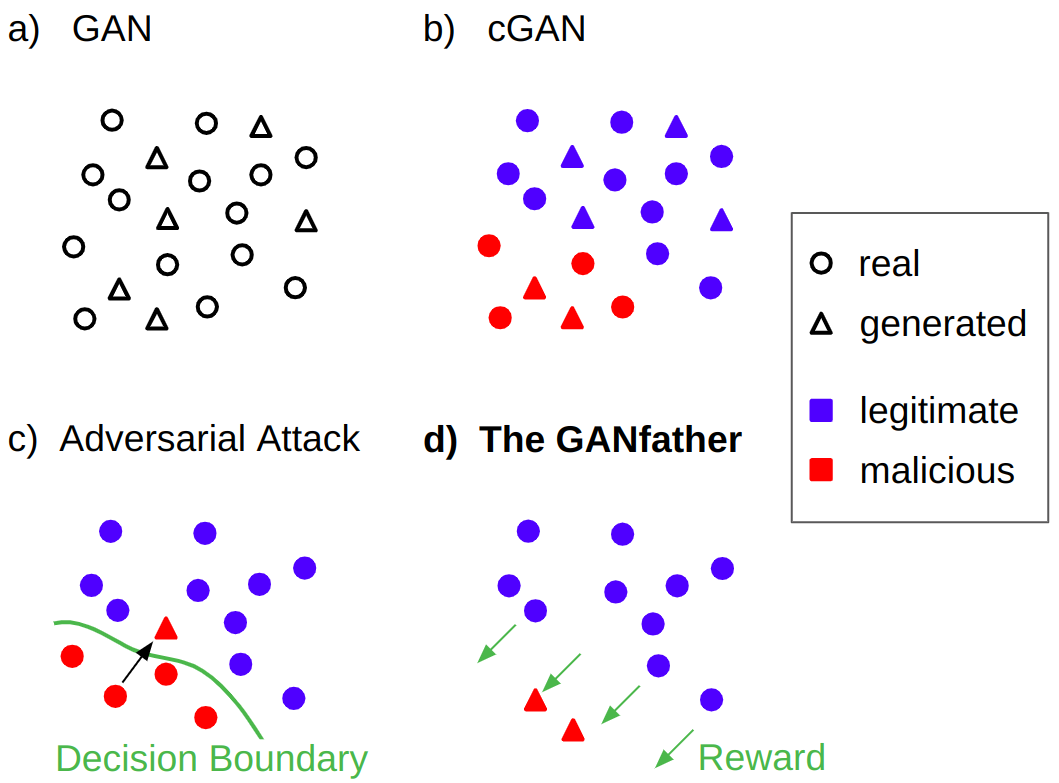}
    \caption{Comparison of our method to some widely used approaches. \textbf{(a)} GAN: a vanilla GAN setup does not require any labels, but one cannot choose the class of a generated sample since the distribution of the data is learned as a whole. \textbf{(b)} Conditional GAN (cGAN): using a cGAN, one learns the class-conditional distributions of the data, allowing the user to choose the class of a generated sample. However, labels are needed to train a cGAN. \textbf{(c)} Adversarial Attack (evasion): starting from a malicious example, perturbations are found such that a trained classifier is fooled and misclassifies the perturbed example. While labels are typically required to select the initial example as well as to train the classifier, eventually the adversarial attacks can be used to obtain a more robust classifier. \textbf{(d)} The GANfather: our method has some desirable properties from the three previous approaches: no labels are needed (as in a GAN), samples of a desired target class are generated (as in a cGAN) and a robust detection system can be trained (as in adversarial training). The combination of these properties in one framework is especially suitable in domains where no labelled data is available.}
    \label{fig:cartoon}
\end{figure}

\section{Introduction}
\label{sec:introduction}

Digital systems' growing dominance in various aspects of our society opens up new opportunities for illicit actors.
For example, digital banking enables clients to open bank accounts more easily but also facilitates complex money laundering schemes. It is estimated that undetected money laundering activities worldwide accumulate to 1.7--4 trillion euros annually \citep{lannoo2021aei}, while operational costs related to anti-money laundering (AML) compliance tasks incurred by financial institutions accumulate to \$37.1 billion~\citep{ray2021celent}. Another example are recommender systems, which are often embedded in digital services to deliver personalised experiences.
However, recommender systems may suffer from injection attacks whenever malicious actors fabricate signals (e.g., clicks, ratings, or reviews) to influence recommendations. These attacks have detrimental effects on the user experience. For example, a one-star decrease in restaurant ratings can lead to a 5 to 9 percent decrease in revenue \citep{luca2011reviews}.

The detection of such malicious attacks is challenging in the following aspects. In many cases, these illicit activities are adversarial in nature, where an attacker and a defence system adapt to each other's behaviour over time. Additionally, labelled datasets are unavailable or incomplete in certain domains due to the absence of natural labels and the cost of manual feedback. For example, besides the large amount of undetected money laundering, the investigation of detected suspicious activity is often far from trivial, resulting in a feedback delay that can last months. 

To address these issues, we propose \textit{The GANfather}, a method to generate examples of illicit activity and train effective detection systems without any labelled examples. Starting from unlabelled data, which we assume to be predominantly legitimate, the proposed method leverages a GAN-like setup~\citep{goodfellow2014generative} to train a generator which learns to create malicious activity, as well as a detection model (the discriminator) learning to distinguish between real data and synthetic malicious data.

To be able to generate samples with malicious properties from legitimate data, we propose to include an additional optimisation objective in the training loss of the generator. This objective is a use-case-specific, user-defined differentiable formulation of the goal of the malicious agents. Furthermore, our method optionally allows to incorporate an existing defence system, as long as a differentiable formulation is possible. In that case, we penalise the generator when triggering existing detection mechanisms. Our method can then actively find liabilities in an existing system while simultaneously training a complementary detection system to protect against such attacks.

Our method has some desirable properties that make it particularly well-suited for adversarial domains where no labelled data is available:

\begin{itemize}[leftmargin=*]
    \item \textbf{No labelled malicious samples are needed}. Here, we assume that our unlabelled data is predominantly of legitimate nature.
    \item \textbf{Samples with features of malicious activity are generated}. The key to generate such samples from legitimate data is to introduce an extra objective function that nudges the generator to produce samples with the required properties. We implicitly assume that malicious activity shares many properties with legitimate behaviour. We justify this assumption since attackers often mimic legitimate activity to some degree, in order to avoid raising suspicion or triggering existing detection systems.
    \item \textbf{A robust detection system is trained}. By training a discriminator to distinguish between the synthetic malicious samples and real data, we conjecture that the defence against a variety of real malicious attacks can be strengthened.
\end{itemize}

While each of these properties can be found separately in other methods, we believe that the combination of all the properties in a single method is novel and useful in the discussed scenarios. In Figure \ref{fig:cartoon}, we illustrate visually how our method distinguishes itself from some well-known approaches. Finally, while we only perform experiments on two use-cases (anti-money laundering and recommender systems) in the following sections, we believe that the suggested approach is applicable in other domains facing similar constraints, i.e., no labelled data and adversarial attacks, subject to domain-specific adaptations.

%% file: 02_method.tex
\begin{figure*}[ht]
\begin{center}
\includegraphics[width=0.65\linewidth]{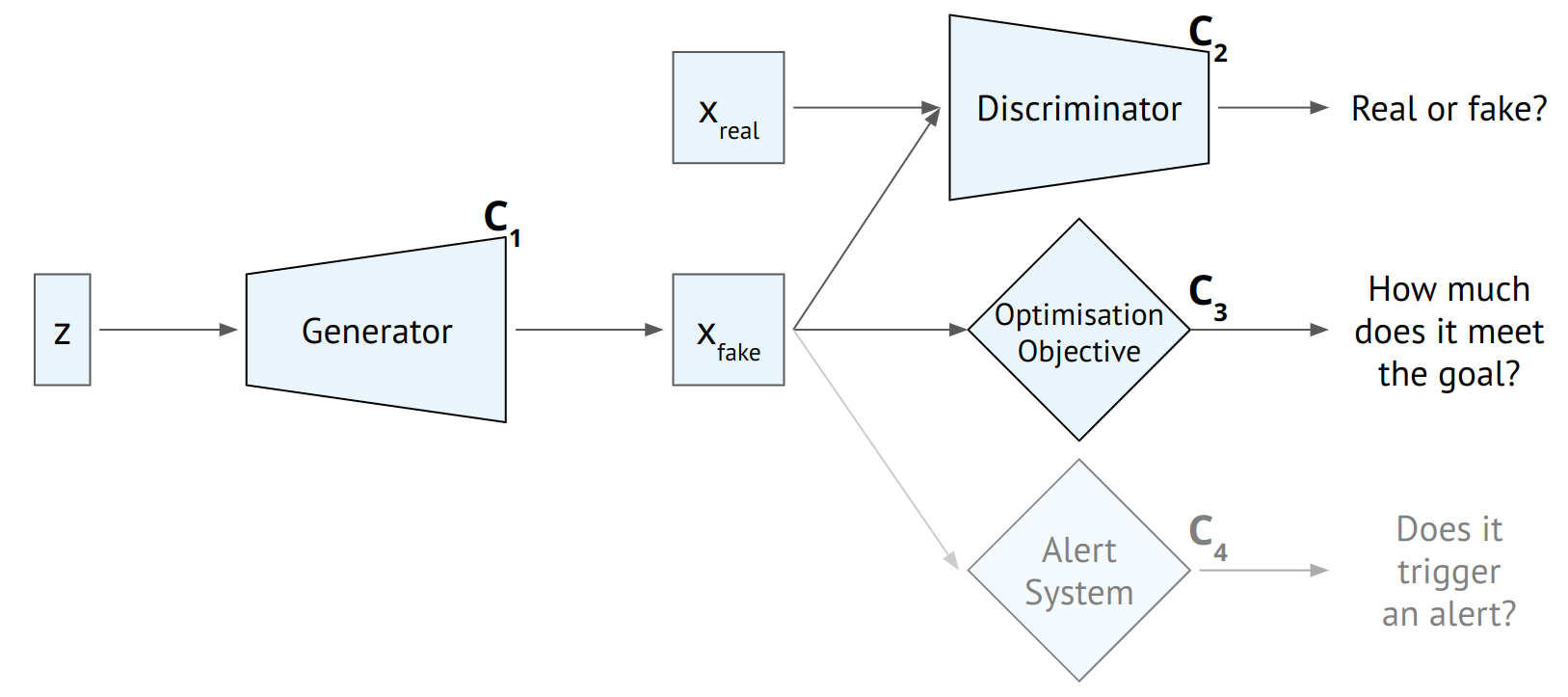}
\end{center}
\caption{\textit{The GANfather} framework. Its main components are a generator, $C_1$, which generates realistic attacks, a discriminator, $C_2$, which detects these attacks, and an optimisation objective, $C_3$, to incentivise the generation of malicious instances. Finally, our method optionally supports the inclusion of an existing alert system, $C_4$.}
\label{fig:03_fulldiagram}
\end{figure*}

\section{Methods}
\label{sec:method}

We provide a general description of our proposed framework in Section~\ref{subsec:overview}.
We proceed to describe two use-cases:  anti-money laundering (AML) (Section~\ref{subsubsec:amlusecase}) and detection of injection attacks in recommendation systems (Section~\ref{subsubsec:rsusecase}).
In Section~\ref{sec:theory}, we show theoretically, in a simplified setting, how our generator's loss function changes the learning dynamics compared to a typical GAN.

\subsection{General description}
\label{subsec:overview}

Figure~\ref{fig:03_fulldiagram} depicts the general structure of our framework.
It comprises a generator, a discriminator, an optimisation objective, and, optionally, an existing alert system. Each component is discussed in more detail below.

\textbf{Generator.} As in the classical GAN architecture, the generator $G$ receives a random noise input vector and outputs an instance of data.
However, unlike classical GANs, the loss of the generator  $\mathcal{L}(G)$ is a linear combination of three components: the optimisation objective for malicious activity $\mathcal{L}_{Obj}(G)$, the GAN loss $\mathcal{L}_{GAN}(G,D)$ that additionally depends on the discriminator $D$, and the loss from an existing detection system $A$, $\mathcal{L}_{Alert}(G,A)$:
\begin{equation}
    \mathcal{L}(G) = \alpha \mathcal{L}_{Obj}(G) + \beta \mathcal{L}_{GAN}(G,D) + \gamma\mathcal{L}_{Alert}(G,A)
    \label{eq:generator_loss}
\end{equation}
where $\alpha$, $\beta$ and $\gamma$ are hyperparameters to tune the strength of each component. The last term is optional, and if no existing detection system is present we simply choose $\gamma = 0$.  Note also that one of the parameters is redundant and we tune only two parameters in our experiments (or one if $\gamma = 0$).

We show in Section~\ref{sec:theory} that the stable point of convergence for the generator in our theoretical example moves away from the data distribution for any $\alpha > 0$. 

\textbf{Discriminator.} The discriminator setup is the same as in a classical GAN. It receives an example and produces a score indicating the likelihood that the example is real or synthetic. Importantly, as explained in Section~\ref{sec:theory}, the generator subject to Equation~\ref{eq:generator_loss} will generate data increasingly out of distribution for larger $\alpha$. Therefore, we do not require the discriminator accuracy to fall to chance level at training convergence, as is usual with GANs. Instead, the discriminator may converge to perfect classification and may be used as a detection system for illicit activity. In our experiments, we use the Wasserstein loss \citep{arjovsky2017wasserstein} as our GAN loss.

\textbf{Malicious optimisation objective.} The optimisation objective quantifies how well the synthetic example is fulfilling the goal of a malicious agent. It can be a mathematical formulation or a differentiable model of the goal. This objective allows the generator to find previously unseen strategies to meet malicious goals.

\textbf{Alert system.} If an existing, differentiable alert system is present, we can add it to our framework to teach the generator to create examples that do not trigger detection (see Equation~\ref{eq:generator_loss}). In that scenario, it is then beneficial for the discriminator to focus on the undetected illicit activity. Whenever the existing system is not differentiable, training a differentiable proxy may be possible.

\textbf{Generator vs. Discriminator views.} If required by the malicious optimisation objective, our generator can be adapted to generate samples which are only partially evaluated by the discriminator. For example, the layering stage of money laundering typically involves moving money through many financial institutions (FIs). However, each detection system operates within single institutions, limiting their view of the entire operation. Our method can be adapted to capture this situation, by generating samples containing various fictitious FIs, but only sending the partial samples corresponding to each FI to the discriminator. In recommender systems, the malicious objective can act on a group of synthetic illicit actors to generate coordinated attacks, while the detection of fraudulent users is typically performed on a single-user level.

\textbf{Architecture optimisations.} In the next sections, we provide more details about the specific architectures used in the two experiments. We note that the architecture details (layer types, widths and number of layers) were first optimised using a vanilla GAN setup (i.e. setting $\alpha=0, \beta=1, \gamma=0$ in Equation~\ref{eq:generator_loss}). With the architecture fixed, the other hyperparameters were tuned as explained in the next sections.

\textbf{Code availability.} The Pytorch code for both models can be found on GitHub
(the link will be provided after double blind review).  

\subsection{Anti-Money Laundering (AML)}
\label{subsubsec:amlusecase}

We tackle the layering stage of money laundering, in which criminals attempt to conceal the origin of the money by moving large amounts across financial institutions through what are known as ``mule accounts''. 

\begin{figure*}
\begin{center}
\includegraphics[width=.9\linewidth]{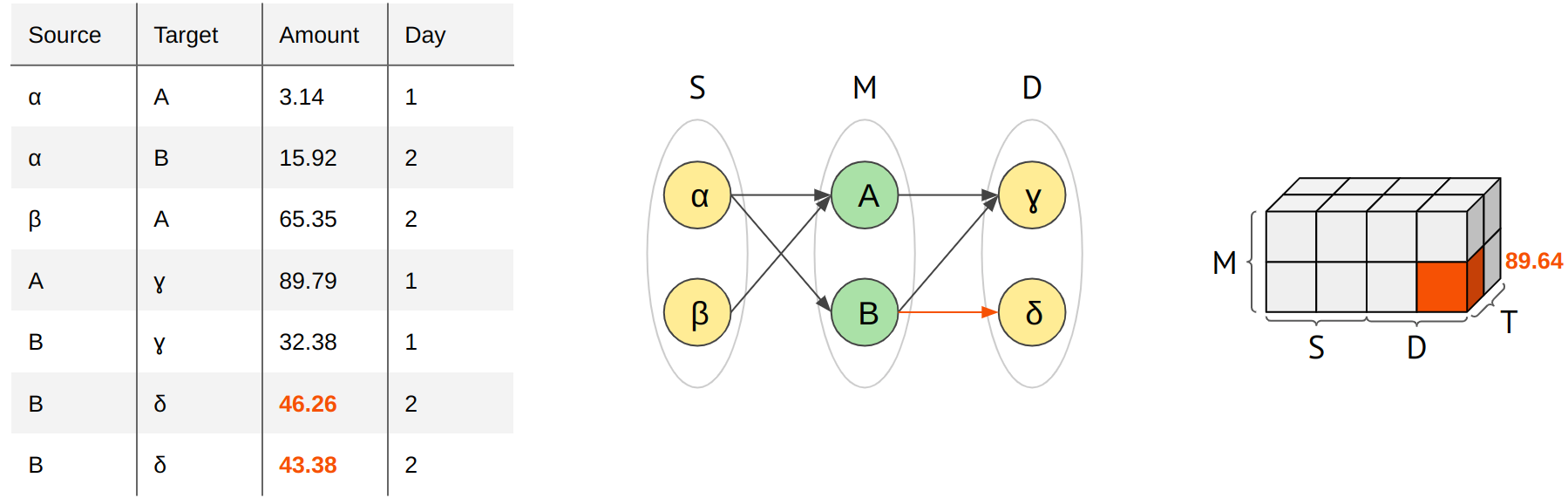}
\end{center}
\caption{Data representation of transactional data. From the raw tabular data, we build the tripartite graph of the transactions, which is in turn represented as a 3D tensor. Here, S stands for Source accounts, M stands for Middle accounts, D stands for Destination accounts and T for the Time dimension.}
\label{fig:03_amldatarepresentation}
\end{figure*}

\textbf{Representing dynamic graphs as tensors.} To represent the dynamic graph of transactions, we can use a 3D tensor as depicted in Figure~\ref{fig:03_amldatarepresentation}. We assume the nodes of the dynamic graph are accounts, and the edges are transactions.
The first two dimensions correspond to the weighted adjacency matrix of the accounts and the third dimension is time.
We discretise the events into time windows of fixed length and group events that belong to the same entry in the tensor by summing their amounts. In other words, each entry $A_{ijk}$ of the tensor corresponds to the cumulative amount sent between account $i$ and account $j$ on timestep $k$.
Our representation covers any dynamic network with a 3D tensor whose size is fixed and pre-specified, which allows us to avoid using recurrent models.
While this approach may limit the size of generated data, domain experts reported that up to 95\% of the money-laundering investigations involve cases containing up to 5 accounts.

\textbf{Architecture.} We implement the generator using a set of dense layers, followed by a set of transposed convolutions. 
Then, we create two branches: one to generate transaction amounts and the other to generate transaction probabilities.
We use the probabilities to perform categorical sampling and generate sparse representations, similar to real transaction data. 
After the sampling step, the two branches are combined by element-wise multiplication, resulting in a final output tensor with the dimensions described above. 

The discriminator receives two tensors with the same shape as inputs: one containing the total amount of money transferred per entry, and the other with the count information (mapping positive amounts to 1 and empty entries to 0). Each tensor passes through convolutional layers, followed by permutation-invariant operations over the internal and external accounts. Then, we concatenate both tensors. We reduce the dimensionality of the resulting vector to a classification outcome using dense layers.

We provide more details about both architectures on our GitHub repository.

\textbf{Money Mule objective.} To characterise the money flow behaviour of layering, where money is moved in and out of accounts while leaving little behind, we define the objective function as the geometric mean of the total amount of incoming ($G(z)_{in}$) and outgoing ($G(z)_{out}$) money per generated account (Equation~\ref{eq:aml_reward}).

\begin{equation}
    \mathcal{L}_{Obj}(G) = -\int \sqrt{ G(z)_{in} \times G(z)_{out} } \cdot p(z) dz
    \label{eq:aml_reward}
\end{equation}
Here $z$ represents random noise input to the generator $G$ and $p(z)$ is its probability distribution. This objective encourages the generator to increase the amount of money sent and received per account and keep these two quantities similar, as observed in mule accounts.

\textbf{Existing Alert System.} In AML, it is common to have rule-based detection systems. In our case, the rules detection system contains five alert scenarios, capturing known suspicious patterns such as a sudden change in behaviour or rapid movements of funds. However, these rules are not differentiable, and our generator requires feedback in the form of a gradient. Hence, we construct a deep learning model as a proxy for the rules system. We hard-code a neural network mimicking the rules' logic operations by choosing the weights, biases and activation functions appropriately. This network gives the same feedback as the rules system would, but in a differentiable way.

\subsection{Recommendation System}
\label{subsubsec:rsusecase}

In this work, we consider collaborative filtering recommender systems. However, our method is compatible with any other differentiable recommender systems.
The system receives a matrix of ratings $R$ with shape $(N_u, N_i)$, where $N_u$ is the number of users and $N_i$ is the number of items.
First, we compute cosine distances between users, resulting in the matrix $D$ of shape $(N_u,N_u)$. Then, we compute the predicted ratings $P$ as a matrix product between $D$ and $R$.
We decided to not represent time since most classical recommender systems do not account for it. However, it is possible to include temporal information using a similar setup to what we described in the AML use case. We also note that, unlike in the AML scenario, we do not have an existing detection system in this setup.
We provide details about the architectures of both the generator and the discriminator on our GitHub repository.

\textbf{Injection Attack Objective.} We define the goal of malicious agents as increasing the frequency of recommendation of a specific item.
The objective function in Equation~\ref{eq:rs_objective} incentivizes the generator to increase the rating of the target item $t$ for every user.

\begin{equation}
    \mathcal{L}_{Obj}(G) = \int \sum_{i}^{N_u}\sum_{j}^{N_i} (P_{ij}(z)-P_{it}(z))_+ \cdot p(z)dz
    \label{eq:rs_objective}
\end{equation}
Here, the matrix of predicted ratings $P$ depends on the random inputs $z$ through the generator $G$ and $(\cdot)_+$ denotes a rectifier setting negative values to zero.

\subsection{Theoretical justification}
\label{sec:theory}
In this section, we provide a theoretical justification to enlighten certain aspects of our setup, in a simplified setting. We will assume no existing detection system is available ($\gamma$ = 0 in Equation~\ref{eq:generator_loss}). In the case such a system would be available, we assume its effect is to limit how far the generated data distribution can be from the real data distribution. Furthermore, we assume that a malicious objective would promote a change in the distribution of at least one feature of the generated data compared to the real data. 

In order to facilitate the analytical calculations, we make the following simplifying assumptions. Firstly, we assume that our data consists of only one feature, for which the regular (legitimate) activity is distributed following a normal distribution $p_{\text{data}}$ with mean $\mu_d$ and standard deviation $\sigma_d$:
\begin{equation}
    p_{\text{data}} = \mathcal{N}\left(\mu_d, \sigma_d\right)
\end{equation}
Secondly, we assume that we do not have any samples of malicious activity but that we know that it is characterised by larger values of this feature compared to the legitimate activity. Thirdly, we assume that the generated data follows a normal distribution $p_{\text{gen}}$ with mean $\mu_g$ and standard deviation $\sigma_g$. Using $\gamma =0$ and $\beta=1-\alpha$ in Equation~\ref{eq:generator_loss}, assuming $0\leq\alpha\leq1$, we can write the training criterion of the generator as:
\begin{equation}
    \mathcal{L}(G) = (1-\alpha) \cdot \left(2 \cdot \text{JSD}\left( p_{\text{data}} | p_{\text{gen}} \right) - \text{log}(4) \right) - \alpha \mu_g \label{eq_loss_gen}
\end{equation}
where the first term denotes the GAN loss \cite{goodfellow2014generative} and the second term denotes our \emph{malicious objective} rewarding the generator to produce samples with properties of the malicious data (i.e. increase the mean $\mu_g$ as much as possible).

We can analytically solve the Jenson-Shannon Divergence (JSD) between the normal distributions, using $\sigma_m^2 = \sigma_d^2 + \sigma_g^2$,
\begin{align}
    \text{JSD}\left( p_{\text{data}} | p_{\text{gen}} \right)
    & = \frac{1}{2} \text{KL}\left(p_{\text{data}} | 0.5*(p_{\text{data}} + p_{\text{gen}}) \right) \nonumber \\ & \qquad + \frac{1}{2} \text{KL}\left(p_{\text{gen}} | 0.5*(p_{\text{data}} + p_{\text{gen}}) \right) \nonumber \\
    & = \frac{1}{2} \left[\log \frac{\sigma_m}{\sigma_d} + \frac{\sigma_d^2 + (\mu_d - 0.5(\mu_d + \mu_g))^2}{2\sigma_m^2} - \frac{1}{2} \right. \nonumber \\  & \left. \qquad + \log \frac{\sigma_m}{\sigma_g} + \frac{\sigma_g^2 + (\mu_g - 0.5(\mu_d + \mu_g))^2}{2\sigma_m^2} - \frac{1}{2} \right]
\end{align}

From this, we can calculate the gradient w.r.t. $\mu_g$:
\begin{align}
    \frac{\partial \text{JSD}(p_{\text{data}} | p_{\text{gen}})}{\partial \mu_g} &= \partial \left( \frac{1}{2} \left[\log \frac{\sigma_m}{\sigma_d} + \frac{\sigma_d^2 + (\mu_d - 0.5(\mu_d + \mu_g))^2}{2\sigma_m^2} - \frac{1}{2} \right. \right. \nonumber \\
    & \left. \left. + \log \frac{\sigma_m}{\sigma_g} + \frac{\sigma_g^2 + (\mu_g - 0.5(\mu_d + \mu_g))^2}{2\sigma_m^2} - \frac{1}{2} \right] \right) / \partial \mu_g \nonumber \\
    & = \frac{\mu_g - \mu_d}{4\sigma_g^2 + 4\sigma_d^2} \label{eq_grad_jsd}
\end{align}

Combining (\ref{eq_loss_gen}) and (\ref{eq_grad_jsd}), we find that the gradient of the training objective of the generator w.r.t. the mean of the generated distribution $\mu_g$ is
\begin{align}
    \frac{\partial \mathcal{L}(G)}{\partial \mu_g} &= \frac{(1-\alpha)}{2} \frac{\mu_g - \mu_d}{\sigma_g^2 + \sigma_d^2} - \alpha
\end{align}

Without loss of generality, we set $\sigma_g^2 + \sigma_{\text{data}}^2 = k/2$, such that
\begin{align}
    \frac{\partial \mathcal{L}(G)}{\partial \mu_g} &= (1-\alpha) \frac{\mu_g - \mu_d}{k} - \alpha
\end{align}

\begin{figure}
    \centering
    \includegraphics[width=0.7\linewidth]{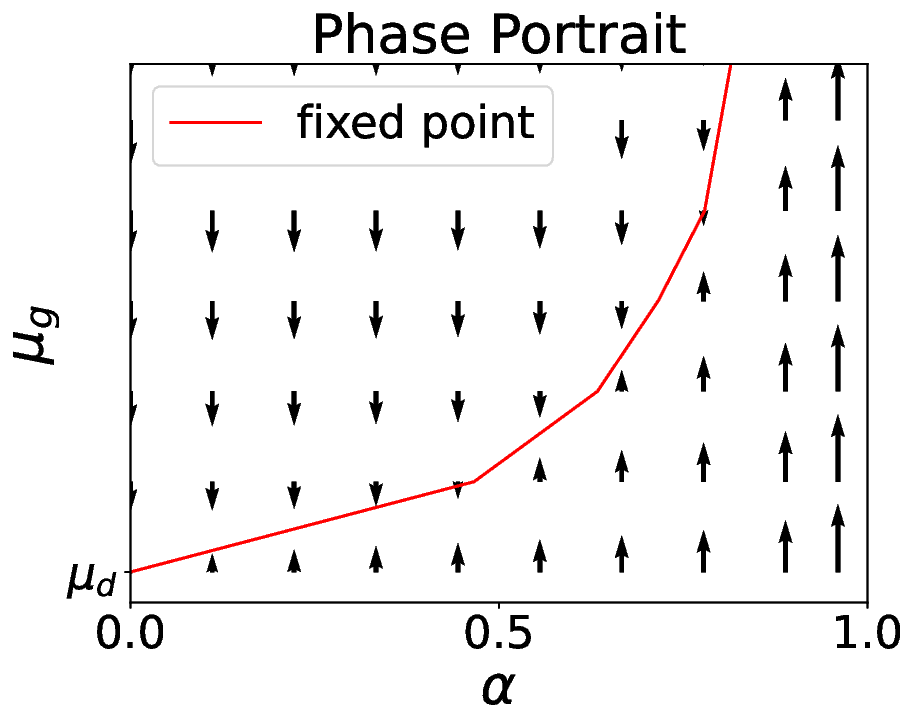}
    \caption{Phase portrait of our toy system. The fixed point of $\mu_g$ depends on hyperparameter $\alpha$. For $\alpha \rightarrow 1$, the fixed point approaches infinity. For $\alpha \rightarrow 0$, the fixed point converges to $\mu_d$. Arrows denote the direction of the gradient $\frac{\partial \mu_g}{\partial t}$.}
    \label{fig:phase}
\end{figure}

Denoting $\frac{\partial \mu_g}{\partial t}$ as the changes of $\mu_g$ over time (i.e. a continuous version of the discrete gradient updates) and $\eta$ as the learning rate, this leads to the following linear dynamical system which we can analyse in function of $\mu_g$, $\mu_{\text{d}}$ and the hyperparameter $\alpha$:
\begin{align}
    \frac{\partial \mu_g}{\partial t} &= - \eta \frac{\partial \mathcal{L}(G)}{\partial \mu_g} \nonumber \\
    &= - \eta (1-\alpha) \frac{\mu_g - \mu_d}{k} + \eta \alpha \nonumber \\
    &= -\eta d\mu_g + \eta d \mu_d+ \eta \alpha
\end{align}
where we defined $d = (1-\alpha)/k$. The stability of this linear system is defined by the sign of $-d$, which is always negative and hence the system has a stable fixed point.
The stable fixed point for this dynamical system is easily found to be 
\begin{align}
    \mu_g^{\star} &= \mu_d + \frac{\alpha}{1-\alpha}k
\end{align}
We plot the phase diagram of the dynamical system in Figure \ref{fig:phase}, showing the fixed point in function of the parameter $\alpha$.

\vspace{2.2cm}

From these simplified setting calculations, we can conclude that:
\begin{itemize}
    \item For $\alpha>0$, our generated data will move away from the real data distribution and increasingly comply with the malicious objective.
    \item Different values of $\alpha$ will result in varying levels of deviation from the real data. In the absence of ground truth to evaluate the system, hyperparameter tuning and empirical testing are necessary.
    \item When generated data deviates from real data, the discriminator will increasingly achieve a perfect performance even at training completion. This is a major difference to standard GAN training.
\end{itemize}

%% file: 03_results.tex
\begin{figure*}
\begin{center}
\includegraphics[width=\linewidth]{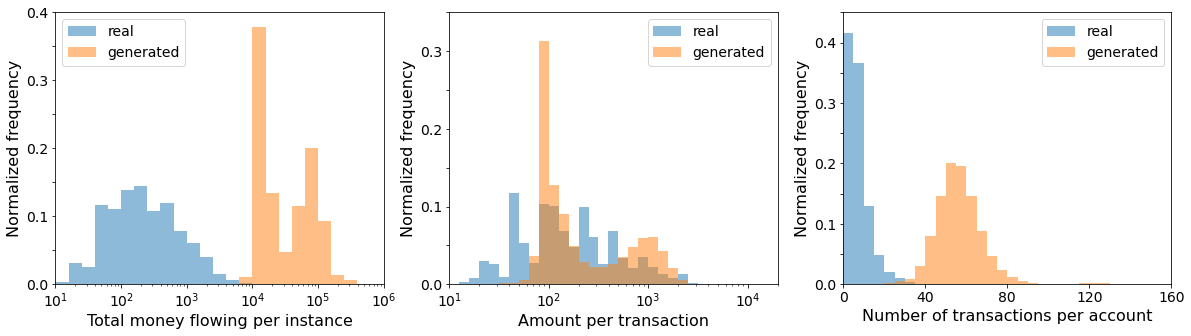}
\end{center}
\caption{Comparing distributions of total money flow, amounts and counts between generated and real data.}
\label{fig:04_aml_comparison_distributions}
\end{figure*}

\section{Results}
\label{sec:results}

We evaluate the efficiency of \textit{The GANfather} to generate and detect attacks in two use-cases: money laundering (Section~\ref{sec:aml_exps}) and recommendation system (Section~\ref{sec:recs_exps}).

\subsection{Money Laundering}
\label{sec:aml_exps}

\textbf{Setup.}
We use a real-world dataset of financial transactions, containing approximately 200,000 transactions, between 100,000 unique accounts, over 10 months\footnote{Due to the confidential nature we cannot disclose the actual dataset.}.
Some of these accounts are labelled as suspicious of money laundering.
We build a real test set of $5000$ accounts, $184$ of which are label positive (suspicious).
We implement \textit{The GANfather}'s generator and discriminator following the architectures presented in Section~\ref{subsubsec:amlusecase}.

\textbf{Results.}
We conduct a hyperparameter random search over the learning rate ($[10^{-4}, 3\times10^{-3}]$) and the weights $\alpha$ (set to $1$), $\beta$ ($[10^2, 10^5]$) and $\gamma$ ($[10^3, 4\times10^3]$) mentioned in Equation~\ref{eq:generator_loss}).

In Figure~\ref{fig:04_aml_comparison_distributions}, we compare the distribution of money flows from such a generator compared to the real data distribution.
We can observe that the generated samples successfully move more money through the accounts than real data (up to 350,000 dollars vs. up to 9,000 dollars respectively).
Interestingly, the distribution of amounts used is similar to real data, and the main difference is the number of transactions used.

Next, we test the detection performance of the trained discriminators on generated data.
To detect potential bias in a discriminator trained solely on samples of the corresponding generator, we first build a \emph{mixed} dataset, where synthetic malicious data is sampled from various generators.
We combine this synthetic dataset with real data, and use it to evaluate the trained discriminators.
Importantly, no retraining on this mixed dataset is performed.
We observe that most discriminators can distinguish between real and generated examples with $100\%$ accuracy, especially those trained with higher values of the $\beta$ hyperparameter (see Equation~\ref{eq:generator_loss}, and note that in this experiment $\alpha$ was fixed to a value of 1).

Then, we evaluate the detection performance on the real test set.
We train a model $DM$ with the same architecture as the discriminator using the mixed dataset mentioned in the previous paragraph.
This training \emph{does not require real labels}, since we use generated data as positive examples (suspicious) and assume that all real examples are negative (legitimate).
After training, we evaluate three detection scenarios: the set of rules mentioned in Section~\ref{subsubsec:amlusecase}; the model $DM$, with the threshold tuned to match the alert rate of the rules\footnote{We assume that the rules are fixed, so we cannot tune the number of their alerts.}; a combination of both (alert if either of them triggers).
The results are shown in Table~\ref{tab:03_aml_reallabels}.
We see that, even though the model $DM$ was trained using only generated data as positive examples, it achieves better detection performance than the rules.
Furthermore, only $10$ of the $128$ alerts of the Rules+Model scenario were alerted by both detection systems, and the true positives had little overlap as well ($5$ out of $54$).
This means that, by including the rules' feedback in the loss of the generator, it learns to create synthetic examples that are not captured by the rules but are similar to real examples of suspicious activity.
As such, a model trained with those synthetic examples can be used to complement the rules, with the advantage of being easy to tune to a desired alert rate.

\begin{table}[H]
    \centering
    \begin{tabular}{l|c|c|c}
        & Alert Rate \% & Recall \% & Precision \% \\
        \hline
        Rules & 1.4 & 13.6 & 36.2\\
        Model & 1.4 & 18.5 & 49.3\\
        Rules + Model & 2.6 & 29.3 & 42.2
    \end{tabular}
    \caption{Detection of real labels.}
    \label{tab:03_aml_reallabels}
\end{table}

\subsection{Recommender System}
\label{sec:recs_exps}

\textbf{Setup.} We use the MovieLens 1M dataset\footnote{https://www.kaggle.com/datasets/odedgolden/movielens-1m-dataset}, comprised of a matrix of $6040$ users and $3706$ movies, with ratings ranging from 1 to 5 \citep{harper2015movielens}.
We implement the generator and discriminator and collaborative filtering recommender system as described in Section~\ref{subsubsec:rsusecase}.
To compute the predicted ratings, during training we take a weighted average of ratings considering all users in the dataset.
We consider all users during training because the initially generated ratings are random, and only providing feedback from the top-N closest users limits the strategies that the generator can learn.
In contrast, we consider the top-400 closest neighbours to compute predicted ratings at inference since we observed empirically that this value produces the lowest recommendation loss.

In this scenario, we do not use an existing detection component, corresponding to $\gamma = 0$ in Equation~\ref{eq:generator_loss}.
We train our networks with 300 synthetic attackers but evaluate the generator's ability to influence the recommender system with injection attacks of various sizes.
We also define four baseline attacks: (1) a rating of 5 for the target movie and 0 otherwise, (2) a rating of 5 for the target movie and $\sim$90 random ratings for randomly chosen movies, (3) a rating of 5 for the target movie and $\sim$90 random ratings for the top 10\% highest rated movies, (4) a rating of 5 for the target movie and $\sim$90 random ratings for the top 10\% most rated movies.

\textbf{Results.} We choose $\beta = 1-\alpha$ in Equation~\ref{eq:generator_loss}, with $0\leq \alpha \leq 1$ and perform a hyperparameter search over $\alpha$.
We observe that increasing $\alpha$ leads to generators whose attacks increasingly recommend the target movie, at the cost of moving further away from the rating distributions of real profiles.

In Table~\ref{tab:03_rs_attack}, we show how many real users have the target movie in their top-10 recommendations, depending on the number of generated users that we inject and how they were generated (through \textit{The GANfather} or the described baselines).
We observe that even with a very limited proportion of generated users (30 among 6040 real users, $~0.5\%$), they are able to greatly influence many real users ($~3.7\%$).
In contrast, the baselines have very small impact on the recommendations of real users.
Lastly, as expected, increasing the number of injected users increases the target movie's recommendation frequency to real users.

\begin{table}
    \centering
    \begin{tabular}{l|c|c|c}
        Generation strategy & 30 users & 60 users & 120 users \\
        \hline
        \textbf{The GANfather} & \textbf{225} & \textbf{290} & \textbf{428} \\
        Baseline 1 & 0 & 0 & 0\\
        Baseline 2 & 0 & 0 & 0\\
        Baseline 3 & 1 & 3 & 7\\
        Baseline 4 & 0 & 0 & 0
    \end{tabular}
    \caption{Number of real users with the target movie in their \mbox{top-10} recommendations, after injecting 30, 60, or 120 \mbox{generated} users.}
    \label{tab:03_rs_attack}
\end{table}

Finally, we analyse the detection of synthetic attacks.
As in the AML scenario we build a test set containing real and synthetic data, where the synthetic data contains a mixture of samples from various trained generators to identify the possible bias of a discriminator to attacks by the corresponding generator.
We then quantify the AUC of the trained discriminators.
We observe that most discriminators trained in a GAN setting (taking turns with a generator to update their weights) achieve around $0.75$ AUC.
Unlike the AML scenario, this suggests that the discriminators are tuned to detect synthetic data from their respective generators, but less so from other generators.
If instead we build a \emph{mixed} training set combining real samples with synthetic data from various generators and use it to retrain a discriminator, it achieves near-perfect classification (above $0.99$ AUC).

%% file: 04_related.tex
\section{Related Work}
\label{sec:related}

\textbf{Controllable data generation.} \citet{wang2022controllable} review controllable data generation with deep learning. Among the presented works, we highlight \cite{de2018molgan}. It leverages a GAN trained with reinforcement learning to generate small molecular graphs with desired properties. Their work is similar to ours in that we both (1) extend a GAN with an extra objective and (2) use similar data representations, namely sparse tensors. However, whereas \cite{de2018molgan} uses a labelled dataset of molecules and their chemical properties, our method does not rely on any labelled data.

\textbf{Adversarial Attacks.} A vast amount of literature exists on the generation of adversarial attacks (see \cite{xu2020adversarial} for a recent review). Such attacks have been studied in various domains and using various setups (e.g. cybersecurity evasion using reinforcement learning \citep{apruzzese2020deep}, intrusion detection evasion using GANs \citep{usama2019generative}, sentence sentiment misclassification using BERT \citep{garg2020bae}). In all cases, a requirement is that labelled examples of malicious attacks exist.

\textbf{Anti-Money Laundering.} Typical anti-money laundering solutions are rule-based~\citep{watkins2003tracking, savage2016detection, weber2018scalable}. However, rules suffer from high false-positive rates, may fail to detect complex schemes, and are costly to maintain. Machine learning-based solutions tackle these problems \citep{chen2018machine}. Given the lack of labelled data, most solutions employ unsupervised methods like clustering \citep{wang2009research, soltani2016new}, and anomaly detection \citep{gao2009application, camino2017finding}. These assume that illicit behaviours are rare and distinguishable, which may not hold whenever money launderers mimic legitimate behaviour. Various supervised methods have been explored \citep{jullum2020detecting, raza2011suspicious, lv2008rbf, tang2005developing, oliveira2021guiltywalker}, but most of these works use synthetic positive examples or incompletely labelled datasets. To avoid this, \citet{lorenz2020machine} propose efficient label collection with active learning. \citet{deng2009active} and ~\citep{charitou2021synthetic} explore data augmentation using conditional GANs. Lastly, \citet{li2020flowscope} and \citet{sun2021cubeflow} propose a metric to detect dense money flows in large transaction graphs, resulting in an anomaly score. Their method does not involve training of a classifier, and instead relies on generating many subsets of nodes and iteratively calculating the anomaly score.

\textbf{Recommender systems (RS) injection attacks.} Most injection attacks on RS are hand-crafted according to simple heuristics. Examples include random and average attacks~\citep{lam2004shilling}, bandwagon attacks~\citep{burke2005limited} and segmented attacks \citep{burke2005segment}. However, these strategies are less effective and easily detectable as most generated rating profiles differ significantly from real data and correlate with each other. \citet{tang2020revisiting} address the optimisation problem of finding the generated profiles that maximise their goals directly through gradient descent and a surrogate RS. Some studies apply GANs to RS to generate attacks and defend the system. \citet{wu2021ready} combines a graph neural network (GNN) with a GAN to generate their attack. The former selects which items to rate, and the latter decides the ratings. \citet{zhang2021attacking} and \citet{lin2022shilling} propose a similar setup to ours in which they train a GAN to generate data and add a loss function to guide the generation of rating profiles. The main differences to our work are the usage of template rating profiles to achieve the desired sparsity, the chosen architecture and loss functions. In our work, sparsity is learned by the generator through the categorical sampling branch (see Section~\ref{sec:method}). Moreover, our method allows the generation of coordinated group attacks by generating multiple attackers from a single noise vector.

%% file: 05_conclusion.tex
\section{Conclusion}
\label{sec:conclusion}

In this work, we propose \emph{The GANfather} to generate data of a class for which no labelled examples are available (malicious activity), while simultaneously training a detection network to classify this class correctly.

We performed experiments in two domains.
In the anti-money laundering setting, the generated attacks are able to move up to 350,000 dollars using just five internal accounts, and without triggering an existing detection system.
Then we show that for a real-world labelled dataset, a model trained with these generated attacks can be used to complement the rules, alerting previously undetected suspicious activity.
In the recommender system setting, we generate attacks that are substantially more successful at recommending the target item than naive baselines.
Then, we train a near-perfect classifier to detect the synthetic malicious activity.
While a real test in a deployment scenario is lacking and should be addressed in future work, we believe our current experiments provide a proof of value of the method.
In these experiments, our method generates a variety of successful attacks, and we therefore believe it can be a valuable method to improve the robustness of defence systems. 

The limitations of our method lie in its assumptions.
Firstly, we assume that the unlabelled data is dominated by legitimate events, and our method would not work in settings where this is not the case.
Secondly, we assume that we can quantify the malicious objective in terms of available features.
In this case, one could argue we can just use the malicious objective as a detection score.
However, the detection system often has a (much) smaller view than the malicious objective.
For example, anti-money laundering systems only view incoming and outgoing transactions for \emph{one} financial institution.
However, our objective can be adapted to generate malicious activity mimicking flows across \emph{multiple} synthetic financial institutions, while keeping the view of the discriminator on an individual institution level.
Thirdly, while our method does not prevent generated data to be very different from real data, we argue that the strength of our method is in generating more subtle attacks that are not immediately distinguishable from real data.
Finally, while we chose the malicious objectives to be as simple as possible in our proof of concept experiments, there is no restriction to make them more complex as long as they are differentiable. 

To conclude, our method fits the adversarial game between criminals and security systems by simulating various meaningful attacks.
If existing defences are in place, our method may learn to avoid them and, eventually, train a complementary model.
Incorporating machine learning models into the detection system typically enhances the detection of illicit activity by triggering more precise alerts, while being easier to fine-tune and maintain.
We believe our work contributes to increase the robustness of detection methods of illicit activity.